# Low-cost Retina-like Robotic Lidars Based on Incommensurable Scanning

Zheng Liu, Fu Zhang and Xiaoping Hong

*Abstract*—High performance lidars are essential in autonomous robots such as self-driving cars, automated ground vehicles and intelligent machines. Traditional mechanical scanning lidars offer superior performance in autonomous vehicles, but the potential mass application is limited by the inherent manufacturing difficulty. We propose a robotic lidar sensor based on incommensurable scanning that allows straightforward mass production and adoption in autonomous robots. Some unique features are additionally permitted by this incommensurable scanning. Similar to the fovea in human retina, this lidar features a peaked central angular density, enabling in applications that prefers eye-like attention. The incommensurable scanning method of this lidar could also provide a much higher resolution than conventional lidars which is beneficial in robotic applications such as sensor calibration. Examples making use of these advantageous features are demonstrated.

## I. INTRODUCTION

Lidar (light detection and ranging) has emerged as an important scientific apparatus in a variety of applications such as environmental survey [1], [2], air aerosol measurement [3], turbulence detection [4] and space exploration [5]. Recent advancements in autonomous robots such as self-driving cars has raised significant demands on smaller and lower-cost lidar sensors, which are indispensable in tasks such as detection, perception and navigation [6]–[8]. Existing robotic lidars [1], although with relatively good performance, can hardly meet the mass-deployment requirements on cost, size and reliability, hindering the development and mass deployment of self-driving cars and other intelligent robots. We propose a new type of lidar sensor that offers advantages in these aspects as well as performance, with potential to accelerate the development of fully autonomous robots. Featuring an angular density distribution similar to human retina, this lidar is ideal for scene perception and tracking applications inspired by the attention mechanism of human vision. The unique scanning method could also enable capturing the field of view (FoV) in high resolution provided enough accumulation time, which is beneficial for various robotic applications such as sensor calibration. This new sensor is also upgradable with straightforward modifications.

## II. RELATED WORK

### A. Conventional mechanical multi-line lidar

The dominant type of existing robotic lidars are based on multi-line mechanical scanning due to its simplicity and the achievements in autonomous driving competitions from DARPA grand to urban challenges [1], [6]–[8]. In these time-of-flight type lidars, multiple pairs of semiconductor transceivers, namely pulsed laser diode (PLD) and avalanche photodetectors (APD), are optically aligned at the focal plane of telescoping lens to form a vertical array of rangefinders. These rangefinders are then rotated along the vertical axis to capture objects along the line-of-sight at full azimuthal angles. As an example, Fig. 1(a) schematically shows the working principle of a conventional 16-line lidar [1]. PLD and APD assemblies are positioned at the focal plane of the telescoping lenses such that the collimated light beam will be parallel to the line intersecting the semiconductor and the center of lens [9]. Each PLD and APD transceiver pair will be responsible for a ranging beam at a different vertical angle. (In Fig. 1(a) the beam from the last of the 16 transceiver pairs is illustrated as an example.) The FoV of this lidar is determined by the vertical space span of the PLD and APD assemblies (Fig. 1(a)) relative to the telescoping lenses. The vertical resolution is thus determined by the number of transceivers within the FoV. However, assembly automation of these lidars becomes difficult when one needs to align the many transceivers with alignment accuracy (for best detection range) and large space span (for large vertical FoV), i.e. a dynamic range problem.

In sensing, display, communication and control systems, dynamic range typically refers to the ratio between the maximal range and minimal attainable value, and is always finite due to physical limitations [10]. For example, in a size measurement system utilizing still camera machine vision, the minimal measurable value is limited by the pixel size, diffraction limit, optical aberration and their combinations, while the largest measurement range is limited by the CMOS sensor size and focal length. In general, dynamic range is difficult to improve for a given physical system and has been the target for many

This work was supported by Livox Technology Ltd. (A DJI incubated company), HKU start-up fund and SUSTech start-up fund.

Z. L. and F. Z. are with the Department of Mechanical Engineering, University of Hong Kong, Hong Kong, People's Republic of China (e-mail: u3007335@connect.hku.hk; fuzhang@hku.hk)

X. H. is with the School of System Design and Intelligent Manufacturing, Southern University of Science and Technology, Shenzhen, People's Republic of China (e-mail: hongxp@sustech.edu.cn)

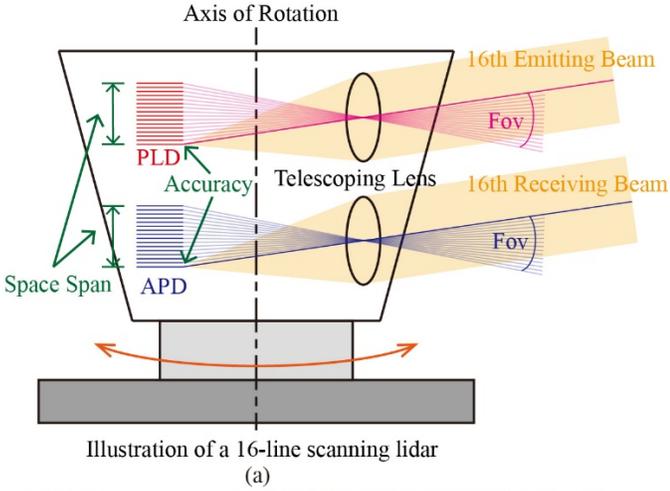

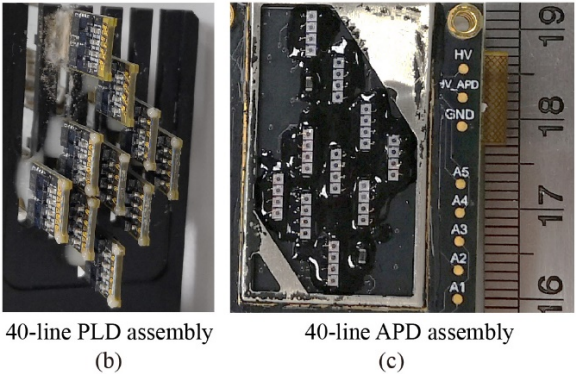

Fig. 1. Conventional multi-line type lidar. (a) Schematics show the working principle of conventional 16-line lidar. PLD and APD assemblies are positioned at the focal plane of the telescoping lenses. The 16th emitting and receiving beams are shown. Note the stringent requirement for accuracy while the alignment is done within a large space span. (b, c) Photo of PLD (b) and APD (c) assemblies from a state-of-the-art conventional 40-line scanning lidar. There are 10 groups of transceiver assemblies, each group having 4 respective PLDs and APDs, to form a 40-line lidar. The PLD assemblies are made by multiple printed circuit boards. These 40-line transceiver elements are located at different vertical positions without an overlap to provide vertical FoV and resolution. Each PLD needs to align accurately with the corresponding APD. Notice the assembly is non-trivial; the PLDs are fixed by glues within the structural fixtures for high precision adjustment. Glues are also used in fixing APDs.

research topics across different fields.

In the alignment of multi-line lidar system, a dynamic range could be defined as the ratio between the space span and the smallest alignment accuracy. As illustrated in Fig. 1(a), the space span is defined as the total vertical space a PLD/APD is allowed to move. The smallest alignment accuracy is defined as the minimum tolerable displacement to best align the PLD and APD so that angularly the emitter beam maximally overlaps with the receiving beam. With the current design of the multi-line lidars, space span is usually on the order of ~5 cm (Fig. 1(b) & 1(c)), and sometimes can go up to more than 10 cm, while the accuracy requirement is on the order of ~10 μm as the active areas of the PLD and APD dies are on the order of ~100 μm. These numbers give a dynamic range above 5000 (5 cm / 10 μm), which is difficult for existing alignment and assembly automation. The limitations could include limited pixel numbers for machine vision, inaccurate rough positioning over large movable scale due to strain or thermal expansion, insensitivity of alignment feedback signals or simply lack of tools to accurately position the PLD and APD within a small housing. Furthermore, these semiconductors need to be glued to the fixtures due to the large flexibility (span) requirements. The complicated gluing process involving initial curing and thermal treatment further complicates the dynamic range problem. (Fig. 1(b) & 1(c)) Therefore, manufacturing of this type of lidar requires lengthy alignment procedures from skilled labors and thus renders prohibitive yield and cost for the automotive market. Additional cost also arises from the many transceiver pairs and their respective high-speed electronics, as vertical resolution is directly related to the number of transceiver pairs.

### B. Risley prism pairs

Risley prism pair, composed of two refractive prisms serially mounted (Fig. 2(a)), provides another simple yet versatile optical scanning method. Unique advantages in prism-based approaches such as low cost, small form factor and robustness permits their wide uses in microscopy [11], interferometry [12], infrared imaging [13], infrared countermeasure [14], explosive detection [15], free space communication [16] and aerial object detection [17]. We propose that these features can be explored in robotic lidar settings, especially to overcome the manufacturing obstacle of existing mechanical lidars and to enable scaling up of autonomous robots such as self-driving cars. In this article, we systematically explore the design and application of low-cost robotic lidars based on the Risley prisms.

The resulting point density distribution is investigated and compared to human retina. Incommensurable scanning, a unique feature associated with Risley prism optical steering, is discussed theoretically and experimentally in terms of robotic applications. We also demonstrate performance upgradability in this design in terms of scanning density increase and customized field of view specifically for autonomous driving. Examples exhibiting the advantages of this type of lidars are provided.

### III. DESIGN OF PRISM-BASED LIDAR

### A. Construction of prism-based lidar

In Fig. 2(a), we present the new lidar based on prism scanning. It consists of two separate modules (transceiver and scanner) that are packed sequentially with a co-axial design. In the transceiver module, the pulsed laser light emitted from the PLD is directed by a tilted mirror, and an aspherical lens is used to collimate the light towards the scanner and the far field. After hitting the object at the far field, the returned signal beam enters the same lens and focuses onto the APD. The distance is measured by calculating the time-of-flight between the emitting and receiving pulses. (Note that the reflector is designed to optimize the overall transceiver efficiency. Due to the difference in numerical apertures of the emitting and receiving beams, this reflector size is carefully chosen to reflect most of the PLD light while still allow the majority of the receiving photons to hit the APD.) A scanner modal, comprising two rotating prisms, is used to direct the transceiver beam to different directions. With this separated prism-based scanner module, we no longer require the scanning function and scanning FoV to be associated with the positioning of the transceivers. Due to this reason, the dynamic-range requirement in the transceiver module is mitigated; only a single alignment of the transceiver packages within a small space span is needed.

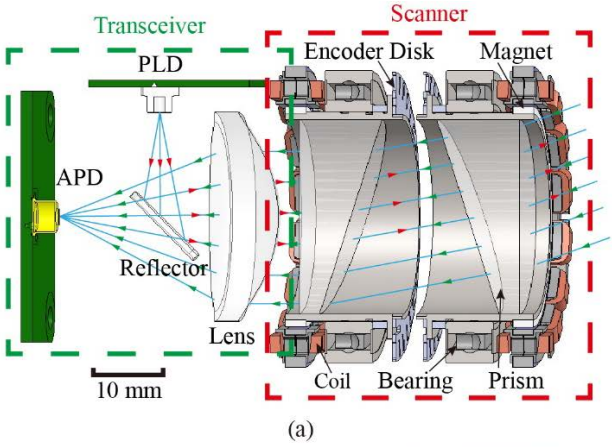

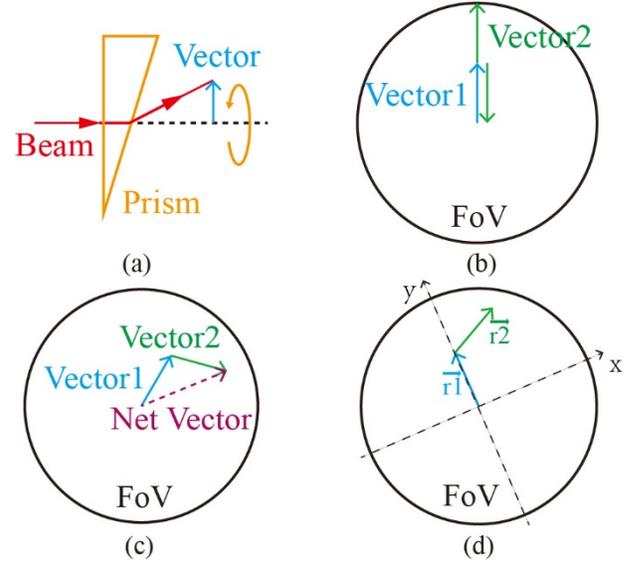

Fig. 3. Scanning principle of prism-based lidar. (a) Illustration of the deflection angle by a single prism in paraxial approximations. The deflection of the optical beam is rotating with the prisms. (b) The parallel and anti-parallel addition of the two vectors from the two prisms define the rim and the center of the field of view (FoV). (c) Access to arbitrary point inside the FoV by vector addition. (d) To calculate the radial density of the point cloud distribution, a local coordinate is used to align with $\vec{r1}$ (first prism vector) for simplicity.

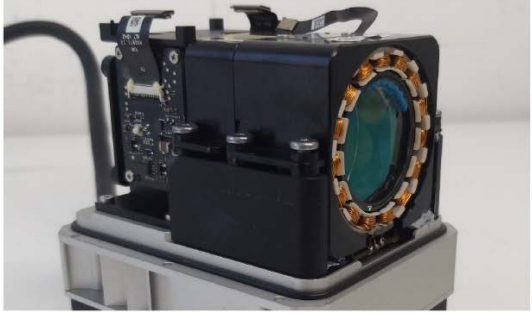

(b)

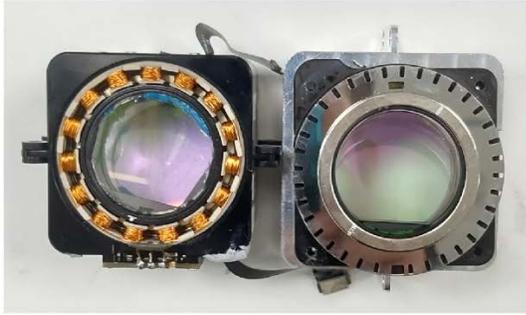

(c)

Fig. 2. Physical structure of the prism-based lidar. (a) A schematic illustration of the design. The transceiver and scanner modules (outlined with dashed boxes) are responsible for range-finding and scanning respectively. The size of the overall lidar is small (a 10 mm scale is indicated). (b) Photo of the assembled lidar (without enclosure box). (c) Photos of the dismantled scanner module comprising motors, prisms and encoder disks.

Fig. 2(b) and 2(c) show the prototype photos of the prism scanner module. It features two oppositely mounted hollow-motors (brushless) that we specifically designed for this lidar. The two prisms, attached with periodically-poled NeFeB magnet rings, are mounted as rotors inside the bearing apertures. Currents in the external stator coils are supplied and controlled by the respective motor drivers to rotate the prisms at desired rotation speeds and phases, by which the beam is scanned.

Although the actual beam angle calculation is done with exact refraction computations, the scanning process can be understood in paraxial approximations [9], [18] for a more intuitive understanding. The monochromatic beam will be refracted by the prism in a linear way regardless of the incident direction, and we could consider the beam direction been deflected by a fixed angular vector for each static prism (Fig. 3(a)). The two prisms are identical in refractive index and wedge angle, meaning the two vectors will have equal magnitude. Rotation of the prisms about the common central axis causes the rotation of these two equal magnitude vectors (Fig. 3(b, c)). The scanning pattern is a result of adding these two asynchronously rotating vectors. As illustrated in Fig. 3(b), when they move opposite to each other, the net vector points to the center of the FoV, while when they move parallel to each other, the net vector lands on the rim of the FoV. Generally, one can access any target vector inside the FoV as shown in Fig. 3(c). Depending on the difference between the two rotation speeds, the beam will scan either in a spiral pattern or a rosette pattern, as shown in Fig. 4(a) and Fig. 4(b) respectively. In fact, the density distribution is a generic feature of this scanning which does not depend on the relative rotation direction, as illustrated analytically in the next session. We use the rosette configuration in our prototype for simplicity.

With carefully engineered magnetic preload and mass balance, each rotor is capable of fast and reliable rotations up to 12000 rpm. While the overall size of the motor/lidar is kept minimal, optical aperture size in this design can be kept as large as possible (Fig. 2) to maximize the receptor signal-to-noise ratio hence the detection range performance. This is another advantage of transmissive prisms instead of 2-axis rotating mirror scanners; the size is significantly reduced for the same optical aperture. With the 905 nm laser satisfying Class I laser safety requirement, the detection distance of our device is 260 meters for an 80% reflectivity object and with false alarm rate below 0.01% under direct sunlight (100 klux). The detection distance can reach 500 meters with a slightly increased form factor. The scanning module also ensures high reliability because the rotating components are optical prisms and passive mechanical attachments without any electrical connections, avoiding prone-to-failure rotating electronics (e.g. slip-rings or wireless transmission setups in multi-line lidars).

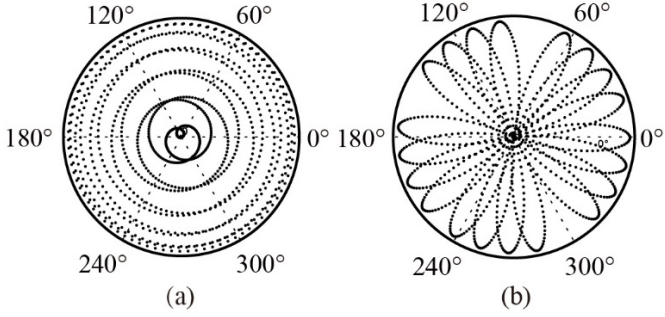

Fig. 4. The spiral and rosette pattern from the lidar for 0.1s, with rotation speeds of small difference (7294 rpm and 6664 rpm, Fig. (a)) and large difference (7294 rpm and -4664 rpm, Fig. (b)) respectively.

### B. Retina-like scanning pattern

Interesting and useful features for autonomous robots can be obtained from the Risley prism lidar. It is noted that the point cloud density distribution becomes retina-like in this lidar. We simulated the angular density distribution of the point cloud from Fig. 4(b) with a reasonable accumulation time (red curve in Fig. 5(a)), and found the density distribution is similar to cone (photoreceptor) distribution in human retina, whose central part (the fovea region) is peaked for sharp vision (the blue curve stands for visual acuity of human eyes in Fig. 5(a)) [19], [20]. This is an interesting feature that permits us to devise eye-like sensing mechanism, as illustrated in the Example Section.

The fovea-like distribution of the scanning density is a generic feature of this scanning method regardless of specific patterns (i.e., rotating speeds) and can be analytically derived. If we scan with relatively long periods, only the radial distribution will be of interests, as the scanning is symmetric in the polar directions. Since the ranging measurement takes place with a constant rate, the density is directly proportional to the duration the pointing vector stays at the radial position. Therefore, in Fig. 3(d) we could specify that the y axis is aligned with the first prism vector, while the second prism vector will be oriented with an angle $(\omega_1 - \omega_2) \cdot t$, which represents the angular displacement of the two prism vectors at time t (Fig. 3(d)). The final vector can be expressed by projecting on to the x and y coordinates.

$$\vec{r} = \vec{r_1} + \vec{r_2}$$
$$\vec{r} = [R(1 + \cos((\omega_1 - \omega_2) \cdot t))]\hat{y} + [R\sin((\omega_1 - \omega_2) \cdot t))]\hat{x} \tag{1}$$

where R is the magnitude of the prism vector. This equation leads to

$$r = \sqrt{2}R\sqrt{1 + \cos((\omega_1 - \omega_2) \cdot t)} \tag{2}$$

The radial density is proportional to the derivative of $r$ w.r.t. time $t$.

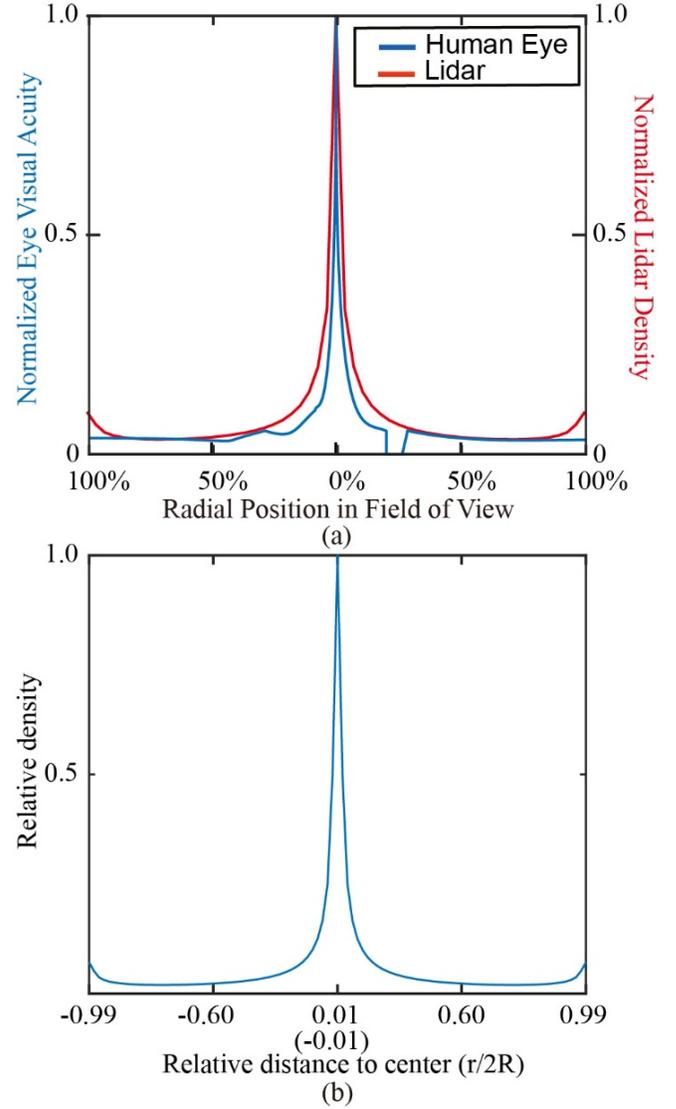

Fig. 5. (a) Comparison of the receptor density distribution of the lidar (red) and human retina (blue). In human retina, the central peaked region is the fovea region and the dip (blind-spot) in retina is due to nerve fibers [19]. (b) The plot of the analytical point cloud density derivation. Notice the divergence at the center and the rim. The distribution agrees with the simulated (a) and experimental distribution (Fig. 14(a)).

$$\frac{dr}{dt} = -\frac{R^2(\omega_1 - \omega_2)}{r}\sin((\omega_1 - \omega_2) \cdot t)$$
$$\sim R(\omega_1 - \omega_2)\sqrt{1 - \frac{r^2}{4R^2}} \tag{3}$$

which is independent of $t$, as we expected for a static distribution. The radial density $\rho$, can be derived,

$$\rho = \frac{N \cdot dt}{2\pi r \cdot dr} = \frac{N}{2\pi R(\omega_1 - \omega_2)} \frac{1}{r\sqrt{1 - \frac{r^2}{4R^2}}} \sim \frac{1}{r\sqrt{1 - \frac{r^2}{4R^2}}} \tag{4}$$

where $N$ is the rate of ranging measurement. The analytical expression of $\rho$ is plotted in Fig. 5(b). The analytical result agrees well with simulation result (Fig. 5(a)) and experimental

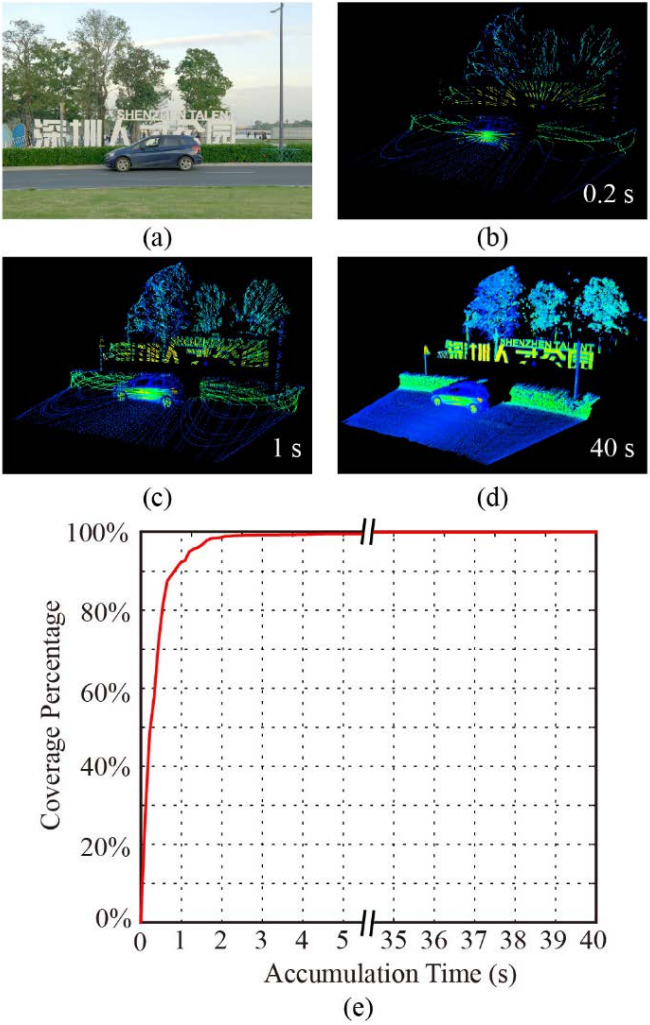

Fig. 6. (a) Photo of a scanned outdoor scenario. (b-d) The scanned point clouds of (a) from lidar with accumulation time 0.2 s, 1 s and 40 s respectively. The color represents the object reflectivity. (e) The FoV coverage percentage as a function of accumulation time.

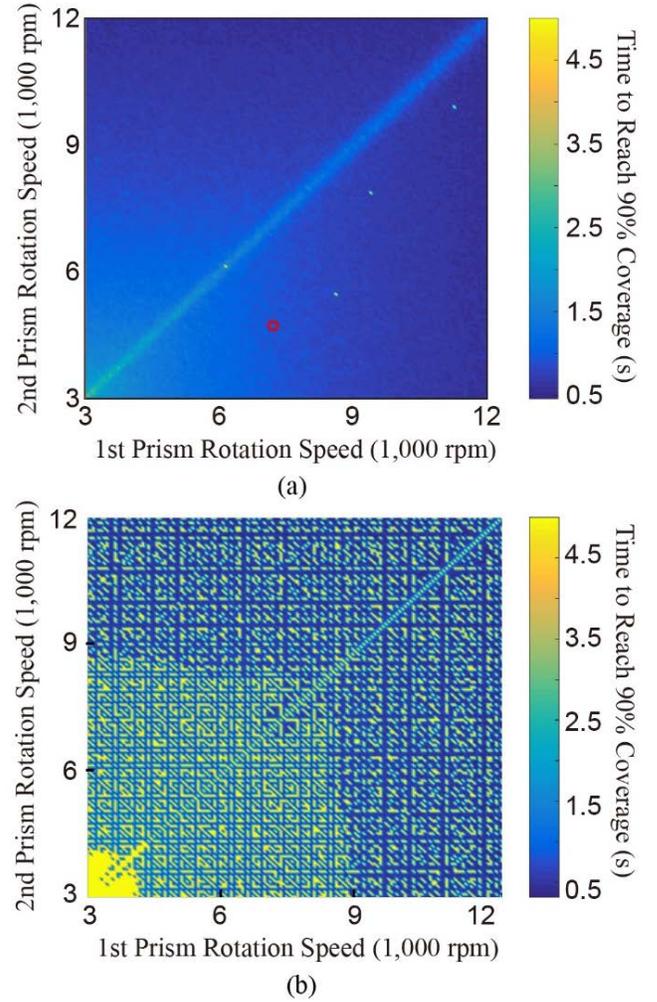

Fig. 7. (a) Simulated time to reach 90% coverage as a function of the two motor speeds (counter-rotating) with 1% noise added in motor speeds. (Configuration of our device is marked with red circle.) This is in great contrast to the case where no noise is present (b). The color scale is capped at 5 seconds. (b) Time to reach 90% coverage with no noise in the motor speeds.

result (Fig. 14(a)).

### C. Incommensurable scanning pattern

Non-repeating pattern is a natural consequence of this lidar design which could provide a high-resolution description of the environment. Fig. 6(a-d) shows the demonstration of this scanning in an outdoor scenario. As the scanning time increases from 0.2s to 40s, the 3D view obtains higher resolution coverage (see definition in Appendix E). Fig. 6(e) shows the percentage of coverage as a function of time, reaching 50% and 90% in the first 0.3 s and 0.8 s respectively and more in longer time. Understanding of this non-repetition behavior is desired. For a repetition to happen the following equations should satisfy simultaneously,

$$\omega_1 \cdot T = n \cdot 2\pi \qquad (5)$$

$$\omega_2 \cdot T = m \cdot 2\pi \qquad (6)$$

where $T$ denotes the time when the first repetition happens, and $\omega_1$, $\omega_2$ are the rotation speeds of the two prisms respectively. Symbols $m$ and $n$ denote integer numbers. By dividing these two equations, we have

$$\frac{\omega_1}{\omega_2} = \frac{n}{m} \qquad (7)$$

meaning the rotation speeds for the two prisms need to be commensurable [21] if any repetition exists, i.e. the ratio is a rational number as denoted by $n/m$. Although the equation is seemingly always satisfied because only rational numbers for $\omega_1$ and $\omega_2$ can be set in the electronic motor control, the repetition rarely happens in realistic cases. It is because the $\omega_1$ and $\omega_2$ would always possess small uncertainties due to the various disturbances in the motor control feedback system (environmental disturbances, sensor noises, control error etc.). Even a relatively small uncertainty is enough to break the commensurability and form a non-repeating pattern. Sometimes active noise injection can be considered if the passive noises are not present in a system. With a typical 1% gaussian noise, we performed a simulation to find out how long the scanning would reach 90% coverage in the FoV for different rotation speeds as shown in Fig. 7(a). Most parameter space colored deep blue in Fig. 7(a) are ideal for non-repeating patterns, in great contrast to commensurable (repeating) situation (Fig. 7(b)) where no noise is present.

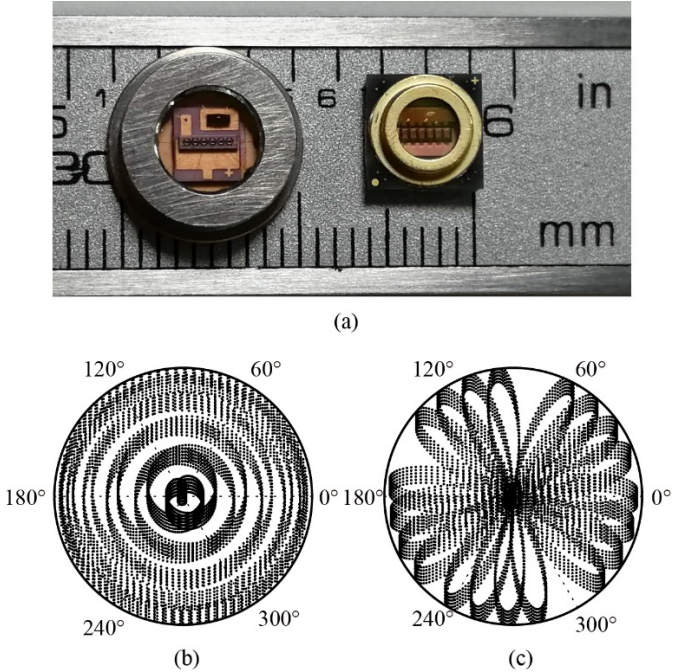

(a)

(b)                           (c)

Fig. 8.   (a) Photo of the packages containing 6 PLD and 6 APD semiconductor dies. (b-c) The spiral and rosette pattern from 6 channels prism-based lidar in 100 ms. Notice the increased density compared to Fig. 4(a) & 4(b).

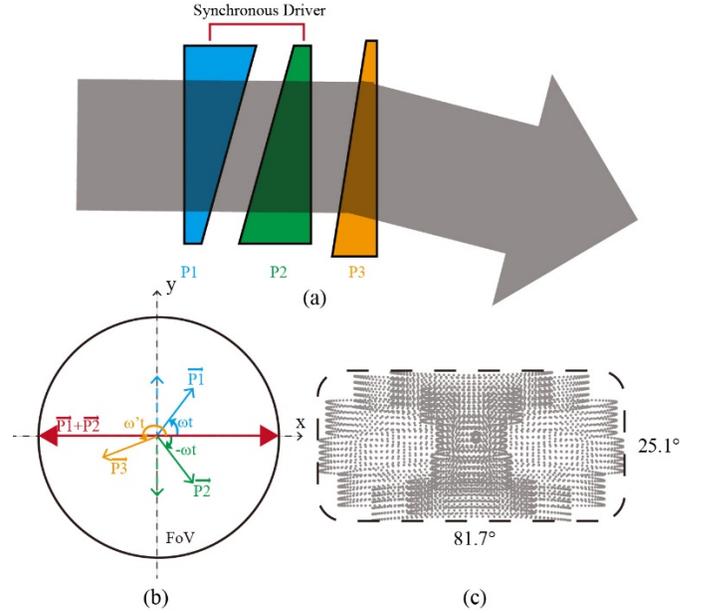

(a)

(b)                           (c)

Fig. 9.   Illustration of close-to-rectangle scanning with three rotating prisms. (A) The configurations of three prisms, where P1 and P2 are controlled synchronously so that their phases are exactly opposite of the other. P3 is driven independently with a smaller deflection magnitude and lower rotation speed. (B) Illustration of the rotation. P1 and P2 are rotated in opposite directions with the same speed. The phase angle difference is controlled accurately by a driver so that the net vector becomes a harmonic oscillator along the x direction. The additional P3 rotates this net vector further to form a close-to-rectangle FoV. (C) The final scanning pattern with this method. Here the 6-element package (Fig. 8(a)) is used for increased density to better meet the requirement of self-driving cars.

## D.  Performance upgradability in point density and FoV

Another advantage of this lidar design is the ease in performance improvement, such as the increased scanning density and the customized FoV. In conventional mechanical lidars, the improvement of both the point density and field of view require placing many additional transceiver pairs, and the difficult arises significantly due to the dynamic range problem as explained in Section II. In the prism-based design, the improvements are much easier thanks to the separation of transceiver module and scanning module.

### 1)  Scanning density improvement with packaged arrays

Instead of using only one PLD and one APD in their respective packages, we could use an array of PLDs and APDs to increase the scanning density. Unlike the difficult manual alignment for the conventional mechanical lidars, packaging technology is readily available to achieve this goal. The new span (size of a transceiver semiconductor package ~ 3 mm) is an order of magnitude smaller than the space span in multi-line design (~5 cm or more). This reduction in span size significantly eases the requirement from dynamic range. Fig. 8(a) shows the actual packages used in a prototype containing semiconductor dies of 6 PLDs and 6 APDs respectively. The semiconductor positioning area has a size about 3 mm, and the assembly process is automated via off-the-shelf commercial packaging equipment. By replacing the single die package in the original design with these array packages, a six-fold increase in point density is immediately available, which significantly reduced the scanning time if a high resolution is needed. We show the spiral and rosette pattern examples from these arrays in Fig. 8(b) and 8(c). One can easily extend to more dies in the packages with high precision, speed and yield. The increased density offers superior density and performance, which is especially useful in self-driving cars. Consequently, significant reduction of cost is expected as compared to the existing multi-line lidars with similar point density.

### 2)  Customized FoV with simultaneously controlled prisms

In many robotic applications, such as high speed self-driving cars, close-to-rectangle FoV is required. In the horizontal direction, a lidar should cover a horizontal FoV as large as possible to sense the surroundings and events. However, vertical FoV does not need to be large because there are not many features on the ground or above in the sky. In this regard, the circular FoV might not be the best choice for this scenario. One should consider increasing the FoV in the horizontal direction while limit the FoV in the vertical direction.

The prism-based approach does provide this capability. A triple prism scanning, as shown in Fig. 9(a), can be designed to achieve this goal. The first two prisms are identical in wedge angle and refractive index and are controlled synchronously so that their rotation angles are exactly opposite to each other. This driving method constrains the net deflecting vector from these first two prisms to be a harmonic oscillator along the x axis (Fig. 9(b)). The third prism is driven independently with smaller magnitude and slower speed to effectively rotates this oscillator vector circularly with lower frequency and contributes to a close-to-rectangle FoV. A scanning example is provided in Fig. 9(c), where the horizontal and vertical FoV are 81.7° and 25.1° respectively. The 6-element package is placed vertically in this example for increased density to better meet the requirement of self-driving cars. This design has led to a product developed by

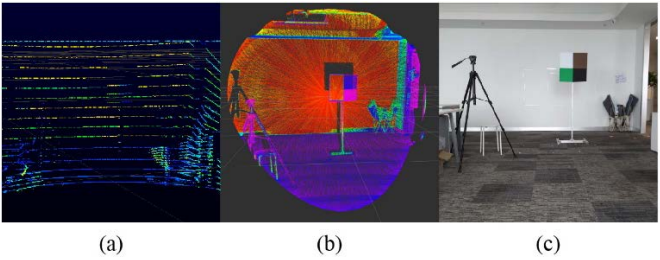

Fig. 10. Sample data of the calibration environment. (a) Data collected from a conventional 16-line mechanical lidar (b) Data collected from our lidar (c) The actual environment.

Livox Technology, known as Horizon[1].

With increased density from packaged multi-transceiver arrays and the novel control of multiple prisms, a versatile method for these high demanding applications is provided with the advantages of low cost, small size, reliability and performance. We present the detailed specs of our devices and illustrate their exemplary applications in key tasks of self-driving cars such as robot navigation, mapping, object detection and tracking in the next section.

## IV. APPLICATIONS

### A. Calibration of lidar and IMU

Sensor calibration is usually a critical step in robotics that a slight mismatch of the coordinates could cause inaccurate or false fusion and undermine the system safety. The incommensurable scan can be valuable in extrinsic calibration between the lidar and other sensors. Conventional multi-line lidar [22] usually leads to inaccurate calibration in vertical directions due to the limited resolution; matching two scans reliably along vertical direction is difficult due to sparsity (Fig. 10(a)). To address this, it is usually required to continuously move the lidar [23]–[26]. This, however, complicates the problem by coupling the estimation of motion that may introduce additional errors such as motion distortion [25], [27]. The incommensurable scanning significantly densifies the point cloud given reasonable accumulation time (Fig. 10(b)), enabling the lidar extrinsic parameters to be accurately calibrated at multiple static poses without motion.

In this experiment, we take the calibration of lidar and IMU (inertial measurement unit) as an example, Fig. 11(a) shows a lidar-IMU sensor set installed on a robotic ground vehicle. The sensor set rotates by roughly 10 degrees and translates 0.10 meters from its origin in all 6 degree of freedom (DOF). In each rotation/translation, the sensor set stays static at that pose for 10 seconds, leading to 99.73% coverage ratio (see Fig. 6(e)). Based on the dense point cloud, the lidar relative transformations are then determined by matching the two respective scans based on the normal distribution transform (NDT) method [32] although a variety of other methods could also be used such as iterative closest point (ICP) matching [28], [29], non-rigid point registration [30], and feature-based registration [33], [34].

On the other hand, the IMU relative transformation are determined by integrating the angular velocity and linear acceleration measured by the gyroscope and accelerometer, respectively. The biases of the IMU are estimated during each static pose, and linearly interpolated during the movement between two consecutive poses. With the relative poses determined by lidar point cloud registration and IMU integration, the determination of extrinsic parameters is a standard hand-eye calibration problem and is solved by methods in [35], [36].

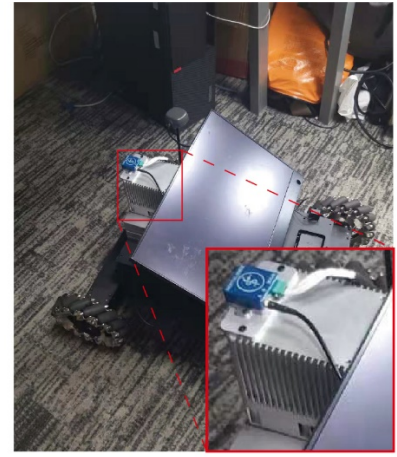

(a)

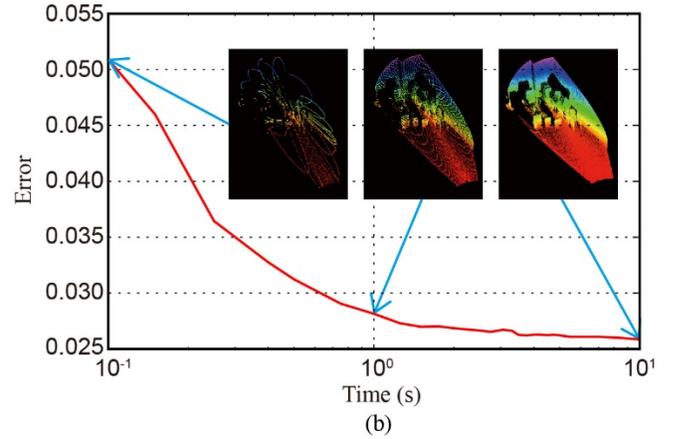

(b)

Fig. 11. Extrinsic calibration of the lidar and an IMU on a robotic ground vehicle. (a) The lidar-IMU setup on a robotic ground vehicle for calibration. The inset magnified the lidar and the IMU setup where they are tightly bound to each other. (b) The calibration error versus the accumulation time at each pose. The inset pictures show the point cloud data at the various accumulation time during the calibration.

We evaluate the calibrated extrinsic by comparing its projected lidar transformation with the ground truth, which are determined by registering the respective point scans accumulated for sufficient long times (e.g., 200s). Assume the calibrated extrinsic parameter is $\overline{X}$, the ground true lidar transformation between two poses is $A$, and the transformation integrated from the IMU data is $B$. Then, the error metric is defined as

$$err = ||Log(A) - Log(\overline{X}B\overline{X}^{-1})|| \quad (8)$$

where Log is the logarithm function on SE(3) and transforms an element in SE(3) to $\mathfrak{se}(3)$ [37].

We calculate the error when the extrinsic parameters are calibrated using different accumulation times and average the error for many different pairs of poses. The results are shown in Fig. 11(b). It is seen that as the accumulation time increases, the calibration error decreases monotonically as expected due to higher lidar resolution (Fig. 11(b), inset pictures). The error drops rapidly during the first second and then decreases slower, in good agreement with the coverage percentage in Fig. 6(e). The improvement brought forth by the incommensurable scanning could be clearly seen from the error reduction. More complicated tasks such as camera-lidar calibration and multi-lidar calibration could also be benefited with the same principle. Generalization to shorter accumulation time or even real time is also feasible [25], [38].

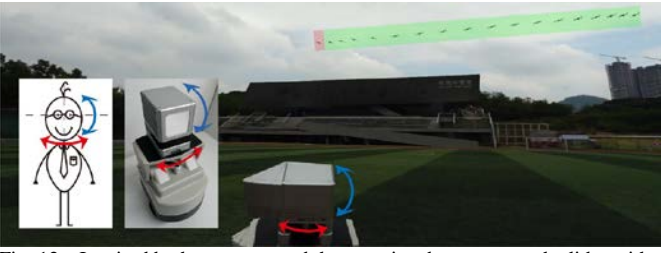

Fig. 12. Inspired by human eye and the associated movement, the lidar with a 2-axis gimbal could perform automatic tracking of the UAV intruder with the central "fovea" region. The pink highlight region is where first detection happens, while the green highlight region denotes the active tracking phase.

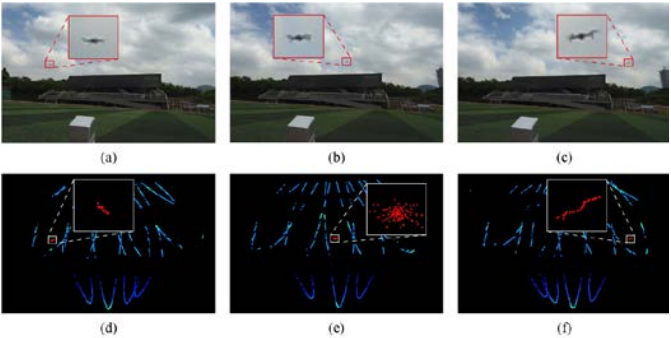

Fig. 13. (a-f) The intruder captured by an RGB camera (a-c, visual guidance only) and detected by the lidar (d-f). The intruder detection on lidar data are computed online automatically. The insets show the magnified pictures for better illustration.

### B. Intruder UAV detection and tracking

Intruder detection and tracking is emerging as an important field in robotics. Multi-line lidars is insufficient in detecting the intruders if they appear in the gap between any two lines. We offer an approach for real-time intruder detection and tracking with this new lidar, whose retina-like resolution and incommensurable scanning provide unique advantages. With the capability to cover the entire FoV, the incommensurable scanning ensures to detect an object, meanwhile the increased point density at the center of the FoV enables accurate tracking. These features are similar to human eyes, where the retina has a central region (i.e. the fovea) with high visual acuity and a surrounding peripheral region that is sensitive to grosser features, especially moving objects [19]. After a successful object detection by the peripheral regions, eyes are turned by extraocular muscles, the neck or the body to project the image of the object onto the high-resolution fovea region and permit its tracking and recognition [20] (inset of Fig. 12). Inspired by this, the lidar can be augmented with a two-degree-of-freedom gimbal system for eye-like robotic object detection and tracking (see Appendix F for details of the developed system). As a proof-of-concept demonstration, a UAV (unmanned aerial vehicle) intruder detection and tracking is considered. We first freeze the gimbal and examine the UAV detection by manually flying the UAV intruder horizontally through the center of the lidar FoV. Throughout the flight, the UAV is being constantly detected (Fig. 13(a-f)), even sometimes with only relatively few points (Fig. 13(d, f)). The number of points detected from the UAV at different locations of the FoV (Fig. 14(a)) agrees well with the lidar point density distribution from theoretical analysis (Fig. 5(a)). To exploit the high density at the lidar "fovea" region and obtain a high definition perception of the object, we actuate the two motors of the gimbal according to the feedback location of the intruder UAV (Fig. 12). With the feedback controller, the lidar quickly tracks the intruder with the "fovea" region once the object appears inside the FoV (Fig. 12, Movie S1). The tracking time and accuracy is shown in Fig. 14(b) and is mainly limited by the gimbal motor performance in our experiments. With the intruder location information from the lidar as well as the gimbal motor angle feedback, the flight trajectory of the intruder can be computed (Fig. 14(c)), useful for further actions from intruder management. Additional cameras (with or without fovea features) could be used in combination with this lidar to enable a more accurate detection and classification of the intruders. In comparison, camera-only systems will not be able to detect the UAV distance and the complete trajectory. The advantages from this bio-inspired robotic sensing system could be of great help for a broad scope of detection and tracking applications such as safety surveillances, industrial monitoring and autonomous driving.

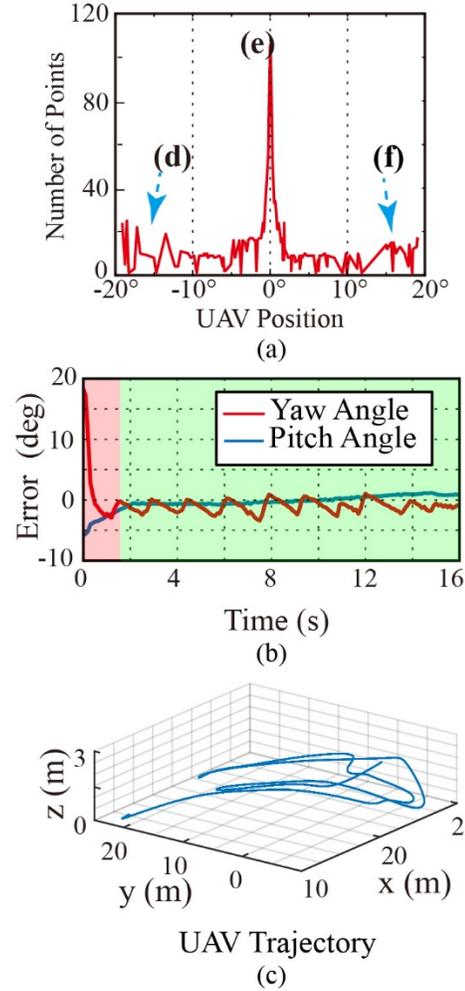

Fig. 14. (a) The number of points detected from the intruder at different locations along the horizontal line which is through the FoV center. (b the angular error from tracking with the same marking color as Fig. 12. (c) The computed flight trajectory of the intruder with lidar and gimbal data.

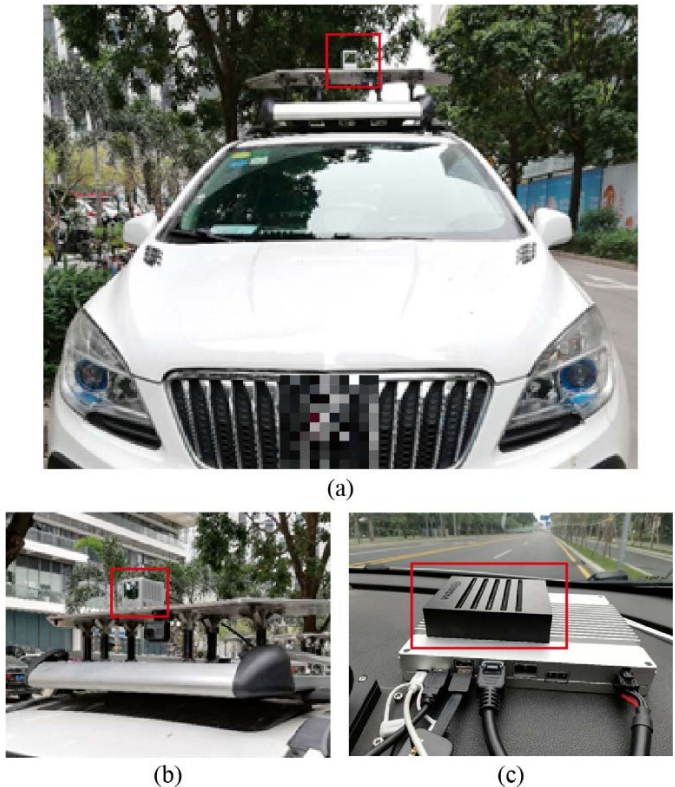

(a)

(b)                              (c)

Fig. 15. Lidar data collection vehicle. (a-b) The Livox Horizon lidar is installed on the vehicle rooftop (c) Nvidia Xavier is used to process the lidar data in real-time. A customized board is developed to power the lidar and route its data to the Xavier for real-time processing.

### C. Demonstration of applications in self-driving cars

In this section, we demonstrate the applicability of our lidars in self-driving cars. We use the upgraded lidar configuration named Horizon as detailed in Section III.D.

Fig. 15(a-c) shows the system configuration consisting of a Horizon lidar for data collection, a Nvidia AGX Xavier[2] for data processing, and a customized board for data routing. We demonstrate two applications which are essential in self-driving cars: object detection and tracking, and lidar odometry and mapping. The algorithm of each application is detailed as below. Both of the two algorithms run on the Xavier in real-time.

#### 1) Object detection and tracking:

The detection program consists of three parts, namely, detection, segmentation and tracking. The detection part uses an end-to-end neural network [39] to extract, classify, and predict the size, location, and orientation of objects of interests, including car, bus, truck, pedestrian, bicycle, and motorbike. The network was trained on a proprietary dataset specifically labeled for the Horizon lidar. The second part, the segmentation, splits the ground points based on the points height and normal vector. Then the ground, foreground objects and background objects are clustered and fused with the detection results in the first part to produce more reliable and accurate segmentation results. Finally, the last part, the tracking, builds on the detection and segmentation results in the previous two parts, pairs for each object in the current frame according to the distance of these objects from the last frame, and smooth the trajectory of each object via a Kalman filtering method. The final results can be seen in the video demonstration at https://youtu.be/sqYGFJVR1HU.

#### 2) Lidar odometry and mapping (LOAM):

We adopt the lidar odometry and mapping algorithm to the Horizon lidar [40]. After receiving a frame (i.e., 100ms) of point cloud, the algorithm extracts edge and plane feature points and register them in a local map as in [41]. Additionally, an IMU is added and calibrated. In the run-time, the IMU data is pre-integrated as in [42] to provide a reliable initial pose estimation for feature point registration. The demonstration of our lidar odometry and mapping in both urban and high-way environments can be seen at https://youtu.be/Aw7I6H7Wj1U.

## V. Conclusion

The prism-based scanning method provides a new machinery in robotic lidar sensors, albeit adoption difficulties could arise from existing algorithms which are designed for conventional multi-line lidars. With simple setup, low cost, low profile and good robustness, we believe this new lidar design will be gradually welcomed by academia and industry, and new autonomous robotic applications will be enabled by the retina-like density distribution and ubiquitous incommensurable scanning.

## Appendix

### A. Product Development

Based on this lidar design, Livox Technology has developed a series of product known as Mid-40[3], Tele-15[4] (Increased density and range), Horizon[5] (Increased horizontal FoV and density), with price tags around or below 1000 USD.

### B. Detailed operating parameters for the lidar prototype

The exemplar scanner is composed of two identical prisms with refractive index of 1.51 and wedge angles 18 degrees. The rotating speeds are 7294 rpm and -4664 rpm respectively. The actual beam pointing direction in our device is computed in real-time by the on-board FPGA from refractions happening at the prism surfaces, whose positions are measured accurately by the encoders. The transceiver operates at a constant measurement rate of 100 kHz ~ 300 kHz, each with the following procedures.

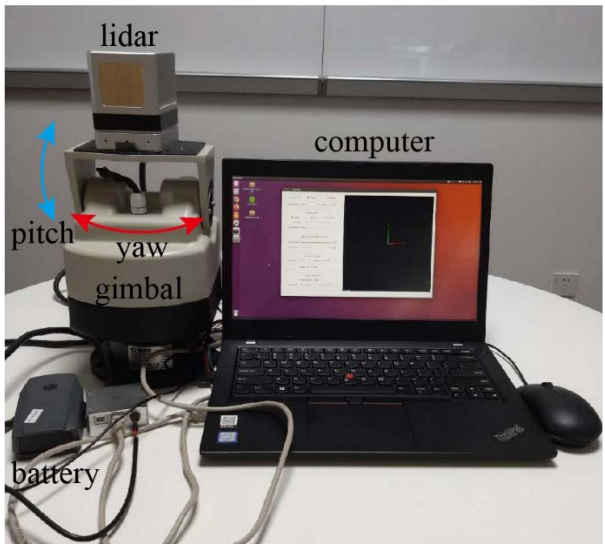

Fig. 16. A close-up view of the lidar gimbal system.

1. Emitter emits a pulse of about 5 ns ~ 10 ns, 50W – 100W peak power.

2. The receiver is triggered and powered on for a few μs.

3. The receiver receives signal if there are returned signals. The internal time-to-digit converter converts the time-of-flight to distance.

#### C. Definition of coverage

To define the LiDAR coverage for a certain time period, the FoV is divided into 100 segments in both horizontal and vertical directions, ideally leading to a total of 10000 voxels. However, due to a circular (not rectangular) FoV and scenario-related constrains, we have 7132 effective voxels. If the laser beam scans to that voxel and a point is collected, the voxel is viewed as filled. Assumed that at time t, n voxels are filled, the formula of coverage percentage is:

$$coverage = \frac{n}{7132} \times 100\% \quad (9)$$

#### D. Intruder UAV detection and tracking

##### 1) System overview

Our gimbal augmented lidar detection system is seen in Fig. 16. The gimbal system is a "PTS-303H" from PTS electronics[6]. It has two-degree-of-freedom: yaw and pitch, respectively driven by two high-torque motors. With the manufacturer supplied software API, the rotation speeds of each motor are independently specified, which enables the gimbal to point along different directions. We developed an integrated software running on a host PC to process all the lidar point cloud data (i.e. for intruder detection), compute and generate motor commands of gimbal systems (i.e. for intruder tracking) and recover the intruder's trajectory in space. The communication between the PC and the lidar is via Ethernet while the communication to the gimbal is via an UART interface.

##### 2) Detection algorithm

**Algorithm 1** summarizes our detection algorithm. It runs in real-time at 10Hz and slices the lidar point stream into a frame (i.e. $P$) for every 100ms. We assume that the space of interest is denoted as $\mathcal{D}$ and it contains no background objects. Then if any point in $P$ lies in $\mathcal{D}$, it is caused by the intruder UAV and should be retained (Line 4-9). The retained points in $P$ are then used to estimate the intruder position by calculating the median coordinate (Line 10). Compared with mean coordinate, the median coordinate we use is more robust to outlier points in $P$.

| **Algorithm 1**: Intruder detection |
|---|
| 1.   **input**: Point cloud $P$ |
| 2.   **output**: Point $p_d$ |
| 3.   **begin** |
| 4.     **for** every point $p_i$ in $P$ do |
| 5.       **if** $p_i \notin \mathcal{D}$ **then** |
| 7.         Delete $p_i$ from $P$; |
| 8.       **end** |
| 9.     **end** |
| 10.     Sort $x$, $y$ and $z$ of points in $P$ respectively and find the median $x_m$, $y_m$ and $z_m$; |
| 11.     $p_d \leftarrow [x_m, \ y_m, \ z_m]^T$; |
| 12.     Return $p_d$; |
| 13.   **end** |

##### 3) Tracking algorithm

**Algorithm 2** summarizes our detection and tracking algorithms. It first runs a detection (Line 3-10) as in **Algorithm 1** to determine the intruder's position relative to the lidar (Line 11-12). If the relative position is below a threshold (e.g. 2°), no action is needed (Line 13-14, Line 18-19). Otherwise, the motor speeds are set proportionally to the relative error (Line 15-17, Line 20-22). Note here $P$ refers to the gimbal's local frame.

| **Algorithm 2**: Intruder tracking |
|---|
| 1.   **input**: Point Cloud $P$ |
| 2.   **output**: motor speeds: $v_y$ (yaw) and $v_p$ (pitch) |
| 3.   **begin** |
| 4.     **for** every point $p_i$ in $P$ do |
| 5.       **if** $p_i \notin \mathcal{D}$ **then** |
| 6.         Delete $p_i$ from $P$; |
| 7.       **end** |
| 8.     **end** |
| 9.     Sort $x$, $y$ and $z$ of points in $P$ respectively and get the median $x_m$, $y_m$ and $z_m$; |
| 10.     $p_d \leftarrow [x_m, \ y_m, \ z_m]^T$; |
| 11.     $e_y \leftarrow \arctan (y_m/x_m)$; |
| 12.     $e_p \leftarrow \arctan (z_m/x_m)$; |
| 13.     **if** $|e_y| < 2°$ **then** |
| 14.       $v_y = 0$; |
| 15.     **else** |
| 16.       $v_y = -k_y e_y$; |
| 17.     **end** |
| 18.     **if** $|e_p| < 2°$ **then** |
| 19.       $v_p = 0$; |
| 20.     **else** |
| 21.       $v_p = -k_p e_p$; |
| 22.     **end** |
| 23.     Return yaw speed $v_y$ and pitch speed $v_p$; |

24. **end**

### 4) Gimbal angle calibration

The detection algorithm in previous sections determines the intruder's position relative to the lidar. To recover the intruder's trajectory in the space (e.g. intruder management), the gimbal orientation is also needed. This is unfortunately not available with the gimbal software API. In our experiments, we calibrate the gimbal's rotation speed by tracking a pre-known feature point (e.g. a room corner) in the lidar point cloud. The calibration builds a lookup table mapping the command to actual rotation speed. Then during the actual intruder tracking, the rotation speed is determined from the command and then integrated to produce the angle estimate.